\documentclass{sig-alternate}\usepackage[]{graphicx}\usepackage[]{color}
\makeatletter
\def\maxwidth{ %
  \ifdim\Gin@nat@width>\linewidth
    \linewidth
  \else
    \Gin@nat@width
  \fi
}
\makeatother

\definecolor{fgcolor}{rgb}{0.345, 0.345, 0.345}

\usepackage{framed}
\makeatletter
 {\par\unskip\endMakeFramed%
 \at@end@of@kframe}
\makeatother

\definecolor{shadecolor}{rgb}{.97, .97, .97}
\definecolor{messagecolor}{rgb}{0, 0, 0}
\definecolor{warningcolor}{rgb}{1, 0, 1}
\definecolor{errorcolor}{rgb}{1, 0, 0}
\newenvironment{knitrout}{}{} 

\usepackage{alltt}

\usepackage{booktabs}
\usepackage{subfig}

\IfFileExists{upquote.sty}{\usepackage{upquote}}{}
\begin{document}

\setcopyright{acmcopyright}

\doi{10.475/123_4}

\isbn{123-4567-24-567/08/06}

\conferenceinfo{KDD '16}{August 13--17, 2016, San Francisco, CA, USA}

\title{Predicting the future relevance of research institutions - The winning solution of the KDD Cup 2016}
\numberofauthors{2} 
\author{
    \alignauthor
    Vlad Sandulescu\\
      \affaddr{Adform}\\
      \affaddr{Copenhagen, Denmark}\\
    \alignauthor
    Mihai Chiru\\
      \affaddr{Bitdevelop}\\
      \affaddr{Stockholm, Sweden}\\
  }
\date{14 August 2016}

\maketitle

\begin{abstract}
The world's collective knowledge is evolving through research and new scientific discoveries. It is becoming increasingly difficult to objectively rank the impact research institutes have on global advancements. However, since the funding, governmental support, staff and students quality all mirror the projected quality of the institution, it becomes essential to measure the affiliation's rating in a transparent and widely accepted way. We propose and investigate several methods to rank affiliations based on the number of their accepted papers at future academic conferences. We carry out our investigation using publicly available datasets such as the Microsoft Academic Graph, a heterogeneous graph which contains various information about academic papers. We analyze several models, starting with a simple probabilities-based method and then gradually expand our training dataset, engineer many more features and use mixed models and gradient boosted decision trees models to improve our predictions.
\end{abstract}


\section{Introduction}

\textbf{The KDD Cup 2016 competition.} The goal of the competition is to rank affiliations, i.e. universities and companies, by the number of their accepted full research papers at top conferences this year. The conferences of interest are SIGIR, SIGMOD, SIGCOMM, KDD, ICML, FSE, MOBICOM and MM and of course it is unknown how many papers an affiliation will get accepted at any of these conferences this year. The organizers allow for any publicly accessible dataset to be used in the challenge. They do however provide a recent snapshot of the Microsoft Academic Graph (MAG) \cite{Sinha:2015:OMA:2740908.2742839}. The competition ran in three stages over the course of a few months. SIGIR, KDD and MM have been chosen to evaluate each team's scores in each phase and the final result is calculated as the weighted sum of these results. The evaluation for each phase is done using the NDCG@20 metric to measure the relevance of the ranking provided by each team: 

\begin{center}
${\mathrm  {DCG_{{n}}}}=\sum _{{i=1}}^{{n}}{\frac  {rel_{{i}}}{\log _{{2}}(i+1)}}$, \\\vspace{3mm}
${\mathrm  {NDCG_{{n}}}}={\frac  {DCG_{{n}}}{IDCG_{{n}}}}$ \vspace{2mm}
\end{center}

where $i$ is the rank of an affiliation, $rel_i$ is the affiliation's relevance score and $IDCG_n$ is the ideal ranking of the affiliations. The evaluation rules are fully described in \cite{RULES}.

\textbf{Our contributions.} We propose and investigate several methods to rank the influence of affiliations at future conferences by predicting their number of accepted full research papers. First, we build a dataset containing records of papers, authors and affiliations using the MAG dataset. We propose three methods to rank affiliations and investigate each of them in the different phases of the competition. Also we show how the accuracy of our predictions increases as we increase the dataset size. We describe thoroughly the feature engineering process, which types of models we used and why and how we tuned our models and feature sets to achieve the best results for each phase.

\textbf{Overall competition results.} We achieve the highest score after all the phases and finish on the first position in the overall results for the entire competition. Moreover we manage to finish on the third position in predicting the ranking of the affiliations participating in the KDD conference this year.

\section{Related Work} \label{related_work}
An approach to measure an author's impact factor (AIF) in the scientific community has been proposed in \cite{Pan2014}. 
They claim many of the metrics used to rank scientific journals do not necessarily reflect the impact of the individual authors, but rather the overall impact of the papers. They extend the impact factor used to rank journals to this new metric which is defined as the average number of citations an author receives in a year for papers published in a fixed time period in the past. They argue this measure can better capture an author's current trend in the academic journals and can help estimate their current rank in the research community. 

Previous work by \cite{feng2014} during the WSDM Cup 2016 mined the MAG and calculated scores for each paper using the number of citations and reference links inside the graph. Affiliation and venue scores have also been calculated based on initial paper scores and refined in an iterative manner.

\section{Dataset and exploration}

\subsection{Dataset}
We use the "2016-02-05" version of the MAG dataset in our research, following the KDD Cup's organizers recommendations. MAG is a large graph containing records of academic papers, citations, authors, affiliations and conference venues. We use the data as provided and we do not attempt to curate the records in any way. First we create a list of the authors from the 2011-2015 selected papers and then extract all the other papers written by these authors since the year 2000. We then traverse the citation graph two steps in breadth-first manner and include these papers as well in our sample. For all sampled papers, we compute the in-degree and the out-degree and use this information to compute paper scores according to the approach described in \cite{feng2014}. We call in-degree the number of papers citing a paper and out-degree, the number of papers found in its reference list. Among other information, the MAG contains the set of keywords listed in each paper. We include in our dataset all the keywords for all sampled papers.

\subsection{Exploratory analysis}
One of the first analysis we carry out is to check for any obvious trends for the accepted papers made by top affiliations to each conference. We consider a top affiliation one which had a large number of accepted papers at a conference in the last five years. The assumption is that for large conferences at least, the top 20 places each year will be taken by more prolific affiliations, likely to have participated in the past to the conference. It is unlikely that affiliations which have few sporadic research papers accepted in the last years are going to be present in the top 20 places. We choose to focus on the first 20 places because the evaluation metric used in the competition is NDCG@20.

In Figure \ref{fig:yearly_papers_counts} we plot the number of full research papers accepted at the KDD conference between 2011 and 2015 for the top 20 affiliations. The length of each line maps the range of the number of accepted papers for the affiliation and the mean number of papers is marked by the larger dot on each line. The plot shows the mean number of papers across all years could be a good predictor to how an affiliation will score in the future.

\begin{knitrout}
\definecolor{shadecolor}{rgb}{0.969, 0.969, 0.969}\color{fgcolor}\begin{figure}[t]
\includegraphics[width=\maxwidth]{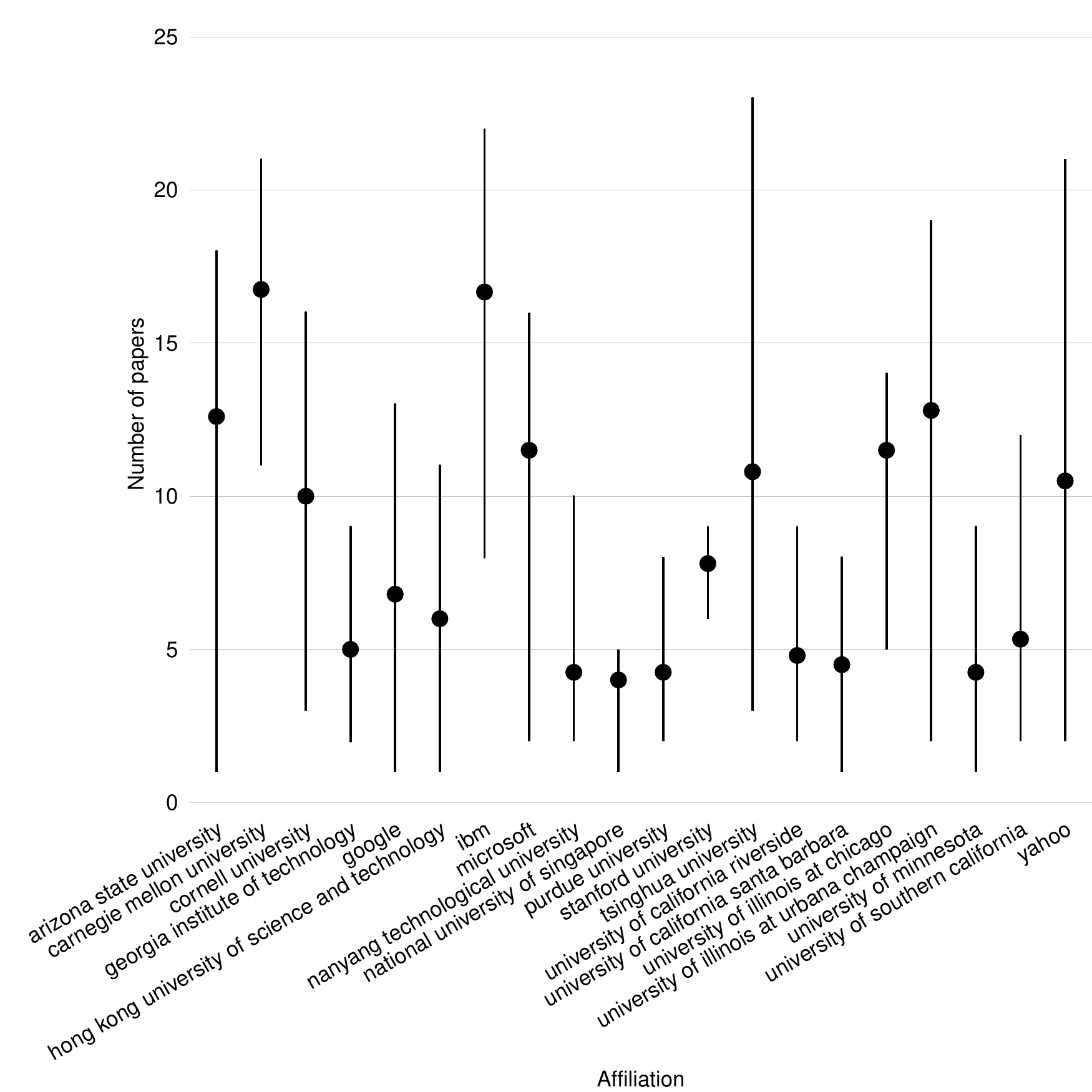} \caption[Full range and mean value of the number of accepted full research papers for top 20 affiliations at KDD between 2011 and 2015]{Full range and mean value of the number of accepted full research papers for top 20 affiliations at KDD between 2011 and 2015}\label{fig:yearly_papers_counts}
\end{figure}

\end{knitrout}

During the second phase of the competition, we try to increase the size of the dataset. For this, we check if the number of full research papers per conference per year is correlated with the total number of papers at the conference for that same year. Besides research papers, at a conference there are usually other types of contributions also present in the MAG such as journal papers, workshop papers or posters. Figure \ref{fig:all_vs_full_counts} shows the two curves look very similar for all the conferences.

\begin{knitrout}
\definecolor{shadecolor}{rgb}{0.969, 0.969, 0.969}\color{fgcolor}\begin{figure}[t]
\includegraphics[width=\maxwidth]{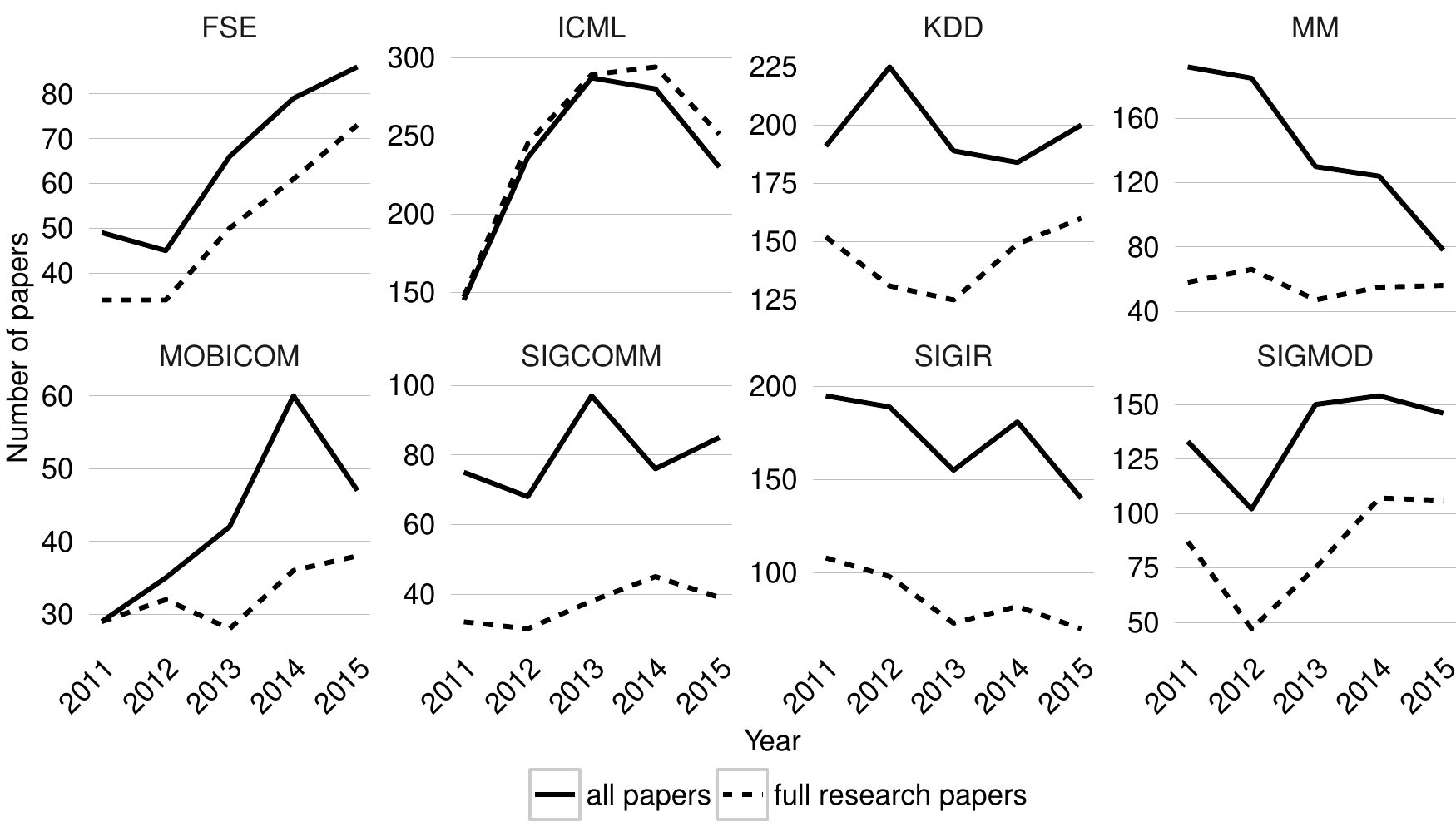} \caption[Number of all the papers vs the full research papers for all the conferences in the competition]{Number of all the papers vs the full research papers for all the conferences in the competition}\label{fig:all_vs_full_counts}
\end{figure}

\end{knitrout}

\section{Models}
In this section, we give an overview of the models we used for each phase. We describe how they evolved over time and how we created the final submissions for each stage.

\subsection{Phase 1}
In the first stage our goal is to set up a good baseline and test how this performs in the competition. Since the public leaderboard is not really informative on a team's performance, we try first to validate our scoring procedure. We compute the probabilities that full research papers belong to affiliations, based on their number of accepted papers across all past five years. The affiliations are ranked according to these probabilities and this represents our submission for the first phase of the competition.

\subsection{Phase 2} \label{models_phase2}
In the second phase and also in the final phase of the competition, we experiment with two classes of models: linear models and gradient boosted decision trees. The former is more interpretable while the latter has more predictive power.

\textbf{Gradient boosting decision trees.}
The gradient boosting decision trees (GBDT) model \cite{trevor2001elements} is effective in dealing with a large number of features and non-linear interactions between the predictor variables and the target variable. The model is a weighted ensemble of weaker decision trees and has a number of advantages which makes it suitable for our purpose. 

We use a boosted tree ensemble which does least-squares residual fitting with a squared-loss objective function. Besides being easy to set up and run, another useful aspect of the GBDT model since its running a greedy feature search, is computing features importance. This adds a degree of interpretability to the predictions and can be used to simplify the model by keeping only the subset of relevant features, which also speeds up the training time. Though, one of the disadvantages of the model is it has a significant of number parameters, which take time to tune in order to avoid overfitting. We overcome this issue by using cross-validation and avoid time-expensive grid searches on the parameters.

\textbf{Mixed models.}
Mixed models \cite{seltman2012experimental} are flexible approaches to deal with correlated measurements which may result for example from experiments on the same group of subjects each receiving the same set of treatments. We can see a connection to the conference-affiliation relation: each affiliation can participate at each of the conferences. There exists an by-affiliation variation and a by-conference variation but also a nested by-affiliation-conference variation. 

It can also be thought as modeling a hierarchical relationship using both fixed and random effects in the same analysis. In our context, we can consider last years relevance scores as being systematic predictors of the relevance this year: these are the fixed effects. Furthermore, we can compute these features for all affiliations. The random effects capture the variability between the different affiliations. Thus, we also account for the nested effect of affiliations for each conference, since we do not want to predict how relevant the affiliations will be in general, but at specific conferences.

\subsubsection{Relevance as target}
Many papers are written by authors affiliated to different universities and/or companies. Our initial models use the number of accepted papers per year per affiliation as main predictor, but this predictor misses the fractional contributions of authors to the final affiliation rank. The relevance score, as calculated in the competition rules, distributes the paper score to each author and thus, each related affiliation receives only a fraction of the score. The final affiliation score is the sum of these fractional contributions. 

Starting with the second phase of the competition, we decide to use the relevance score as the prediction target in our models. We believe this correlates better with the affiliation ranks since the competition's evaluation metric, NDCG@20 is also calculated using the relevance scores. We build the dataset as combinations of (conferences x affiliations x years). The large number of combinations helps grow the dataset from several hundreds samples to several thousands.

\subsubsection{Feature engineering} \label{phase2_feature_engineering}
We already know the mean historical relevance has a strong impact on how relevant an affiliation will be at a particular conference in the future. Besides this, other features such as weighted trends, which are derived from past relevance values, as well as authors publishing trends also have an impact on the future relevance. 

First we compute simple statistics to measure different properties of past years' relevance scores. Then we try to capture the relevance trend across past years. Finally, another set of features is built using the AIF measure. We experiment with the paper scores which are computed using the approach described in \cite{feng2014}. However we do not include them in any of the features used in the competition because our predictions are worse than without them.

\textbf{Statistics-based features.}
Basic statistics of past relevance scores across all years: standard deviation, sum, minimum, maximum, median and mean.

\textbf{Trend-based features.}
Weighted moving-average of relevance scores from past years, using higher weights for recent years and lower weights for years further away in the past. The weights are normalized (sum to 1) and decrease linearly with past years. These trend features are created using weighted averages for windows of two, three and four years.

\textbf{AIF-based features.}
Basic statistics of the AIF computed across all authors from each pair affiliation-conference and across all the years: standard deviation, sum, minimum, maximum, median and mean of the AIF metric.

\subsubsection{Extending the dataset} Figure \ref{fig:all_vs_full_counts} shows that the number of all papers, regardless of their type follows the number of full research papers for all the conferences in the competition. Based on this we extend the dataset even more, using records from all the papers available in MAG for the selected conferences since 2011.

\subsection{Phase 3}
In the third phase, we extend the dataset by adding samples from related conferences and extending the years range. We tune the models to find the best features and number of related conferences to use for the rankings. Also we find there exists a match between the important features found by the GBDT model and the statistically significant features found by the mixed effects model.

\subsubsection{Related conferences}
Most researchers publish their work at different conferences. However they specialize in a specific area and so the conferences they publish at have to be more or less similar at least in a few respects. We use authors and keywords from the papers in MAG to cluster similar conferences together. It is a straightforward way to grow the dataset even more. The intuition behind this is the information from related conferences will enforce the patterns discovered by the models, because prolific affiliations are prolific across all conferences they submit to, not just at one of them. Our assumption is that the same set of features should work better for similar conferences than for dissimilar ones.

We compute the Jaccard similarity for both authors and keywords for any pair of conferences in the MAG. From this, we can determine which conferences are for example most similar to KDD in terms of common authors and common papers' keywords.

Table \ref{tab:kdd_related} shows the most related conferences to KDD.
\begin{table}[ht]
\centering
\begin{tabular}{ll}
  \toprule
By authors & By keywords \\ 
  \midrule
ICDM & CIKM \\ 
  CIKM & ICDM \\ 
  WWW & WWW \\ 
  AAAI & SIGIR \\ 
  ICML & SIGMOD \\ 
  SDM & ICML \\ 
  PAKDD & AAAI \\ 
  ICDE & NIPS \\ 
   \bottomrule
\end{tabular}
\caption{Conferences related to KDD} 
\label{tab:kdd_related}
\end{table}

\subsubsection{Feature engineering}
In this phase, we keep most of the features we used in the previous competition phase, but we try to refine them.

Besides the simple statistics which measure an affiliation's past relevance at a particular conference across all the years, we create four window versions of each of them for the last year up to four years back. We replace the features based on weighted trends from the previous phase with a set of more accurate time series based trend measures.

\textbf{Statistics-based features.}
Basic statistics of past relevance scores across time windows from the last year up to the last four years: standard deviation, sum, minimum, maximum, median and mean.

\textbf{Trend-based features.}
Drift trend of historical relevance scores, which captures the increase or decrease of the relevance over time according to the average change in the past samples. This gives us an estimate of the relevance for the current year which we wrap in a new feature. Also we create five more features based on simple exponential smoothing and we experiment with different smoothing parameters. The new features are all one-step-ahead forecasts, again based on all past relevance scores. Depending on the smoothing parameter, we give exponentially less weight to older observations, making newer observations more important and vice versa. Yet another new feature finds the smoothing parameter value to best fit the data points.

\subsubsection{Extending the dataset even more} \label{phase3_extension}
As we show in Figure \ref{fig:all_vs_full_counts}, the number of full research papers is correlated with the total number of papers an affiliation has at a conference. Although the full research papers are not explicitly marked in the MAG before 2011, we assume this also applies to papers before this year. So we extend our dataset using papers from year 2000 and onward.

\subsubsection{Model tuning}
We observe the GBDT model generally outperforms the mixed effects model on the same set of features. Although it is not always the case it also outperforms the simple model using probabilities. We notice the feature importance matrix, a byproduct of the GBDT model significantly changes when we experiment with different numbers of related conferences. Also there is not a single set of features which performs the best for all the conferences in the final competition phase. We want to be able to train a model with data up to two years and validate against last year's known true relevance scores, or train it with data up to three years ago and test it on true rankings from two years ago and so on.

We aim to find a good configuration of features and number of related conferences for which our GBDT model to always outperform the probabilities model.

\subsubsection{Mixed effects model}
Our mixed effects model contains both fixed and random effects. The fixed effects are the statistics and trend-based features we test the previous models on also. The random effects part is represented by the nested effect of conference-affiliation. The approach makes it possible to observe the importance of each of the explanatory variable and whether their effect on the target variable is significant. 

We try to improve the model by performing backward elimination of non-significant effects. However we do not notice improved accuracy over the GBDT model in general. 

We tested the feature sets used for the best scoring GBDT models with this model and although the accuracy of this model was not as high, it marked as statistically significant all the important GBDT features. This supports and encourages our final submission for this stage of the competition.

\section{Experiments}
In this section we first give specific details on how the size of our dataset size evolves throughout the competition. Then we present our results for all the phases, underlying the final features and model parameters we use to generate our submissions.

\subsection{Data Set}
Table \ref{tab:dataset_stats} shows the dataset size (in number of rows) in every phase of the competition. The model for the first stage uses the smallest dataset of only around 1.3K samples. In the final phase we extend the dataset significantly by increasing the number of related conferences we take into account. We train the final models for the MM conference on nearly 93K samples.

\begin{table}[ht]
\centering
\begin{tabular}{ll}
  \toprule
Item & Sample size \\ 
  \midrule
Full research papers & 3677 \\ 
  Phase 1: probabilities & 1296 \\ 
  Phase 2: full research papers & 8605 \\ 
  Phase 2: all papers & 10900 \\ 
  Phase 3: FSE + 5 related confs & 25136 \\ 
  Phase 3: MOBICOM + 5 related confs & 21872 \\ 
  Phase 3: MM + 10 related confs & 92672 \\ 
   \bottomrule
\end{tabular}
\caption{Dataset description and how it evolves throughout the competition} 
\label{tab:dataset_stats}
\end{table}

\subsection{Phase 1 results}
We observe and show in Table \ref{tab:phase1_results} that predictions made using probabilities computed on more years (e.g. compute the probabilities over the entire range 2011-2014) are generally more accurate. They also follow an ascending trend for SIGIR when predicting the next year (e.g. predict 2015 rankings), than using 2011-2013 to compute the probabilities and predicting 2014 rankings. The predictions for 2013 using only the past couple of years are the worst across all the conferences. For our final submission, we compute the probabilities using the numbers of full research papers between 2011-2015.

\begin{table}[ht]
\centering
\begin{tabular}{lrrr}
  \toprule
Conference & 2015 & 2014 & 2013 \\ 
  \midrule
SIGIR & 0.95 & 0.94 & 0.89 \\ 
  SIGMOD & 0.87 & 0.94 & 0.82 \\ 
  SIGCOMM & 0.93 & 0.95 & 0.77 \\ 
   \bottomrule
\end{tabular}
\caption{NDCG@20 results for the probabilities model in phase 1 for 2013, 2014 and 2015} 
\label{tab:phase1_results}
\end{table}

\subsection{Phase 2 results}
In this phase, we train the models on data from all accepted papers between 2011 and 2015. Table \ref{tab:phase2_features} describes the features used.
\begin{table}[ht]
\centering
\begin{tabular}{lp{0.35\textwidth}}
  \toprule
Feature & Description \\ 
  \midrule
$s(rel)$ & Stats of all previous relevance scores (std, sum, mean, median, min, max) \\ 
  $pw_y(rel)$ & Previous relevance scores in windows from previous year up to $y$ years ago \\ 
  $wt_y(rel)$ & Weighted moving-average of previous relevance scores in windows from previous year up to $y$ years ago \\ 
  $s_{aif}$ & Stats of AIF metrics (std, sum, mean, median, min, max) \\ 
   \bottomrule
\end{tabular}
\caption{Features used in phase 2} 
\label{tab:phase2_features}
\end{table}

The results we obtain from our experiments with records from the full research papers and from all papers are shown in Table \ref{tab:phase2_results}. We also validate our predictions on the full research papers list accepted at SIGIR 2016, which was published on the conference website during the second phase of the competition. Using data from all the papers during the training phase produces far better results than just from the full research papers. The predictions we make for SIGIR 2015 represent the only case where this does not happen. However we find more encouraging the results for this year's SIGIR are more accurate than for the previous year.

\begin{table}[ht]
\centering
\begin{tabular}{lrrrr}
  \toprule
Conference & 2015(F) & 2016(F) & 2015(A) & 2016(A) \\ 
  \midrule
KDD & 0.84 & - & 0.91 & - \\ 
  ICML & 0.87 & - & 0.93 & - \\ 
  SIGIR & 0.92 & 0.93 & 0.87 & 0.97 \\ 
  SIGMOD & 0.84 & - & 0.93 & - \\ 
  SIGCOMM & 0.87 & - & 0.90 & - \\ 
  MOBICOM & 0.70 & - & 0.70 & - \\ 
  FSE & 0.77 & - & 0.80 & - \\ 
  MM & 0.89 & - & 0.94 & - \\ 
   \bottomrule
\end{tabular}
\caption{NDCG@20 results for the GBDT model in phase 2, using all papers (A) and full research papers (F) for 2015 and 2016} 
\label{tab:phase2_results}
\end{table}

\subsection{Phase 3 results}
We describe the features we use in the stage of the competition in Table \ref{tab:phase3_features}.

\begin{table}[ht]
\centering
\begin{tabular}{lp{0.35\textwidth}}
  \toprule
Feature & Description \\ 
  \midrule
$s_y(rel)$ & Stats of all previous relevance scores (std, sum, mean, median, min, max) computed in windows from previous year up to $y$ years ago \\ 
  $sw_y(rel)$ & Stats of previous relevance scores (std, sum, mean, median, min, max) computed in windows from previous year up to $y$ years ago \\ 
  $w_y(rel)$ & Previous relevance scores computed in windows from previous year up to $y$ years ago \\ 
  $dt(rel)$ & Drift trend of previous relevance scores \\ 
  $es(rel)$ & Exponential weighted moving average of previous relevance scores with estimated smoothing parameter \\ 
  $es_\alpha(rel)$ & Exponential weighted moving average of previous relevance scores, computed with a fixed smoothing parameter $\alpha$ \\ 
   \bottomrule
\end{tabular}
\caption{Features used in phase 3} 
\label{tab:phase3_features}
\end{table}

Most of the used features in this phase have been tested in the previous phase of the competition. As described in Section \ref{phase3_extension}, we extend the dataset to contain the last 15 years of papers' records, so we use data from all accepted papers between 2000 and 2015. The number of features after creating all the time-window-based ones is around 50. 

We test the correlation between the features and the target variable, the relevance. Figure \ref{fig:features_correlations} shows the correlation matrices for all the remaining features for KDD and MM after eliminating the AIF-based ones. For KDD all the features strongly correlate with the relevance, while MM shows at most a moderate correlation. 

\begin{knitrout}
\definecolor{shadecolor}{rgb}{0.969, 0.969, 0.969}\color{fgcolor}\begin{figure*}
\subfloat[KDD and 5 related conferences between 2000-2015\label{fig:features_correlations1}]{\includegraphics[width=.49\linewidth]{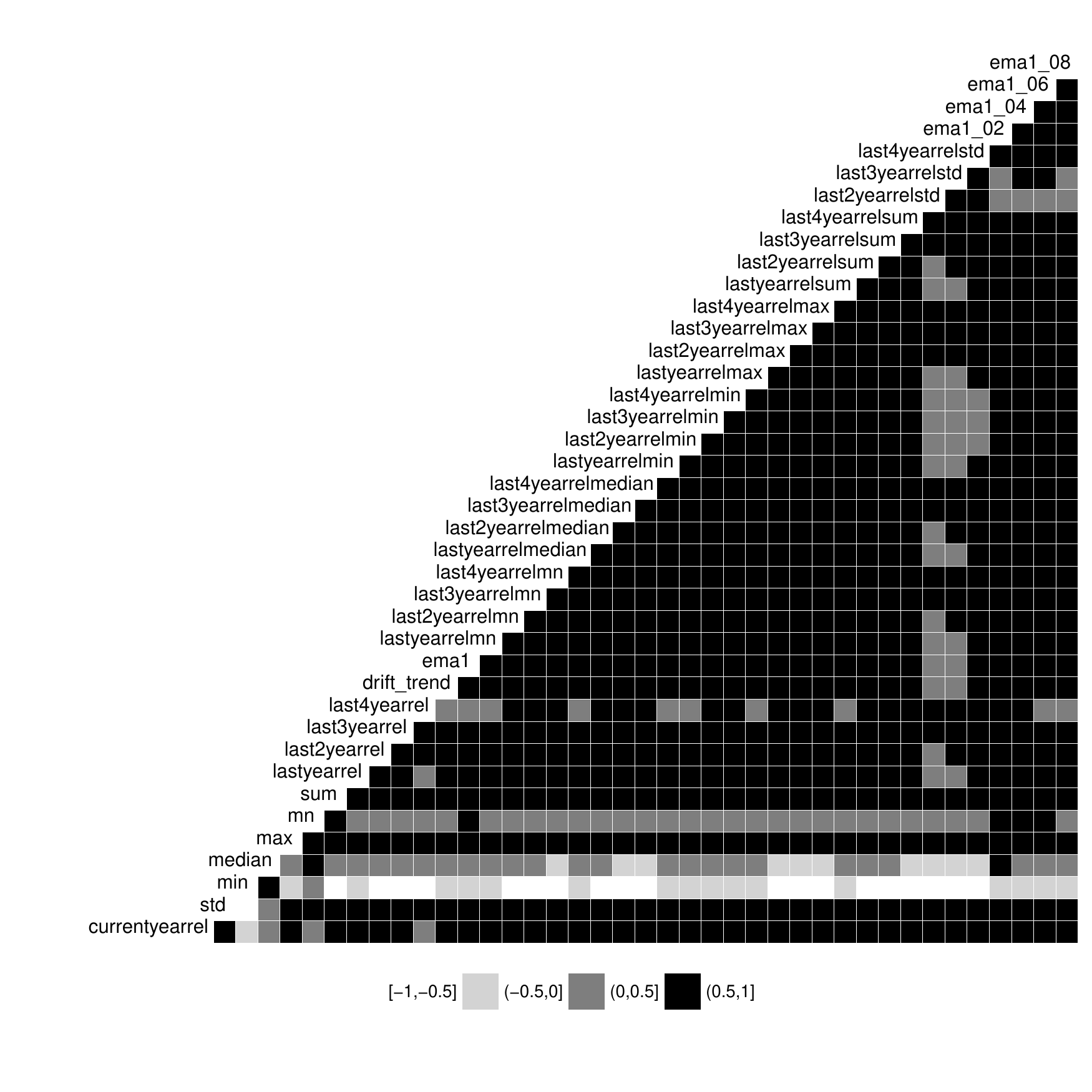} }
\subfloat[MM and 10 related conferences between 2000-2015\label{fig:features_correlations2}]{\includegraphics[width=.49\linewidth]{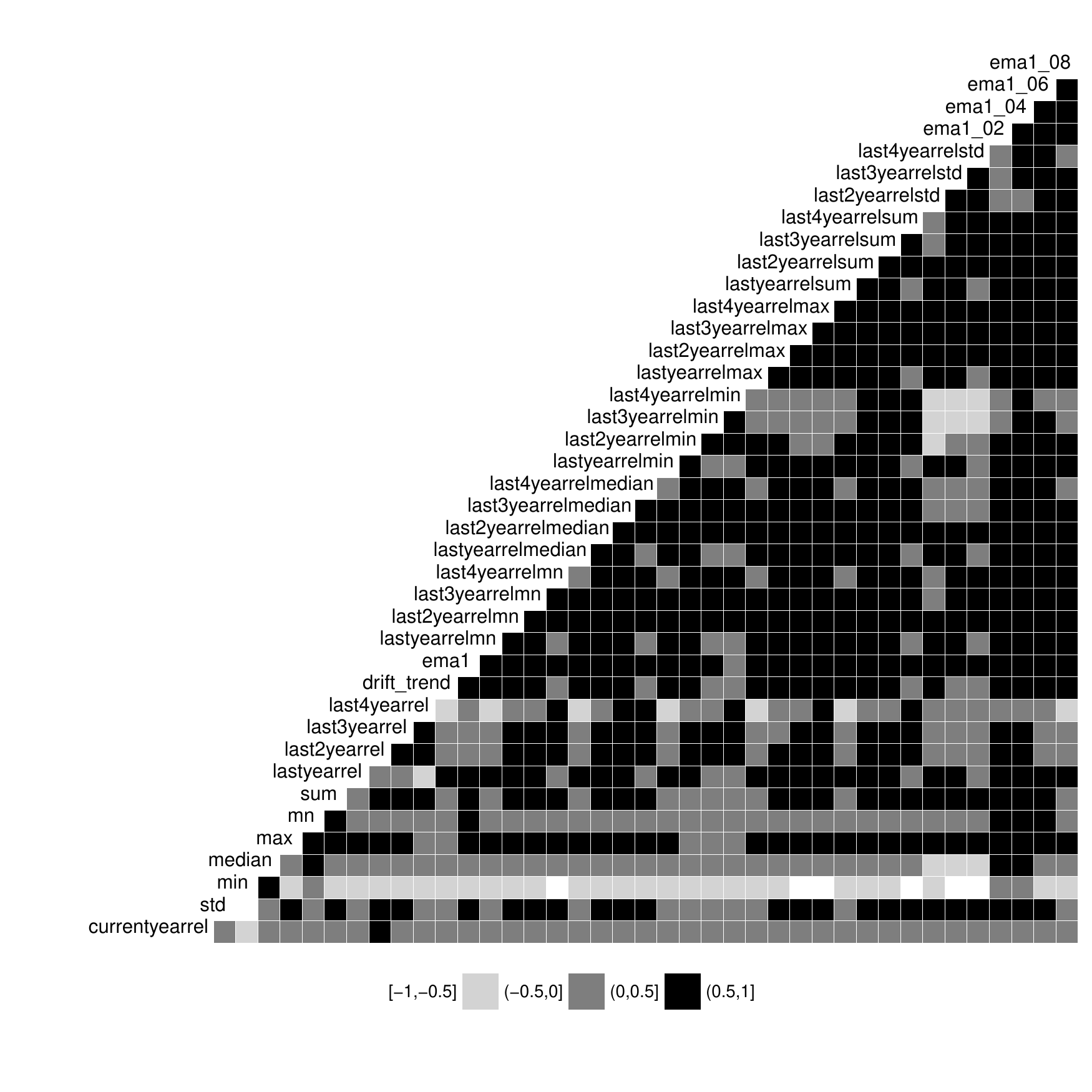} }\caption[Correlation matrices for the features used in phase 3]{Correlation matrices for the features used in phase 3}\label{fig:features_correlations}
\end{figure*}

\end{knitrout}

We search for the features configuration for which the GBDT model gives the best predictions for the conferences in the final phase. First, we create a feature set for the two estimated trend features $dt(rel)$ + $es(rel)$. Next, we compute all the combinations of:
\begin{itemize}
\item the features in Table \ref{tab:phase3_features}, but $dt(rel)$ and $es(rel)$ is replaced with the feature set $dt(rel)$ + $es(rel)$
\item the number of related conferences in the set \{0, 5, 10, 15, 20\}
\item the years between 2012-2015
\end{itemize}

From all the configurations we tested in phase 3 of the competition we choose for each of the conferences the one which outperforms the probabilities model across all years. Table \ref{tab:phase3_results} shows the best features and number of related conferences resulted from the tuning process.

\begin{table}[ht]
\centering
\begin{tabular}{lp{0.25\textwidth}r}
  \toprule
Conference & Features & Related \\ 
  \midrule
FSE & $sw_y(rel)$ + $dt(rel)$ + $es(rel)$ & 5 \\ 
  MOBICOM & $w_y(rel)$ + $sw_y(rel)$ + $dt(rel)$ + $es(rel)$ & 5 \\ 
  MM & $w_y(rel)$ + $dt(rel)$ + $es(rel)$ & 10 \\ 
   \bottomrule
\end{tabular}
\caption{Best feature sets and related conferences number configuration for phase 3} 
\label{tab:phase3_results}
\end{table}

Figure \ref{fig:benchmark_simple_probabilities} shows the corresponding results of the best features configurations for each conference. All the scores of the GBDT model are above the probabilities model baseline.

\begin{knitrout}
\definecolor{shadecolor}{rgb}{0.969, 0.969, 0.969}\color{fgcolor}\begin{figure}
\includegraphics[width=\maxwidth]{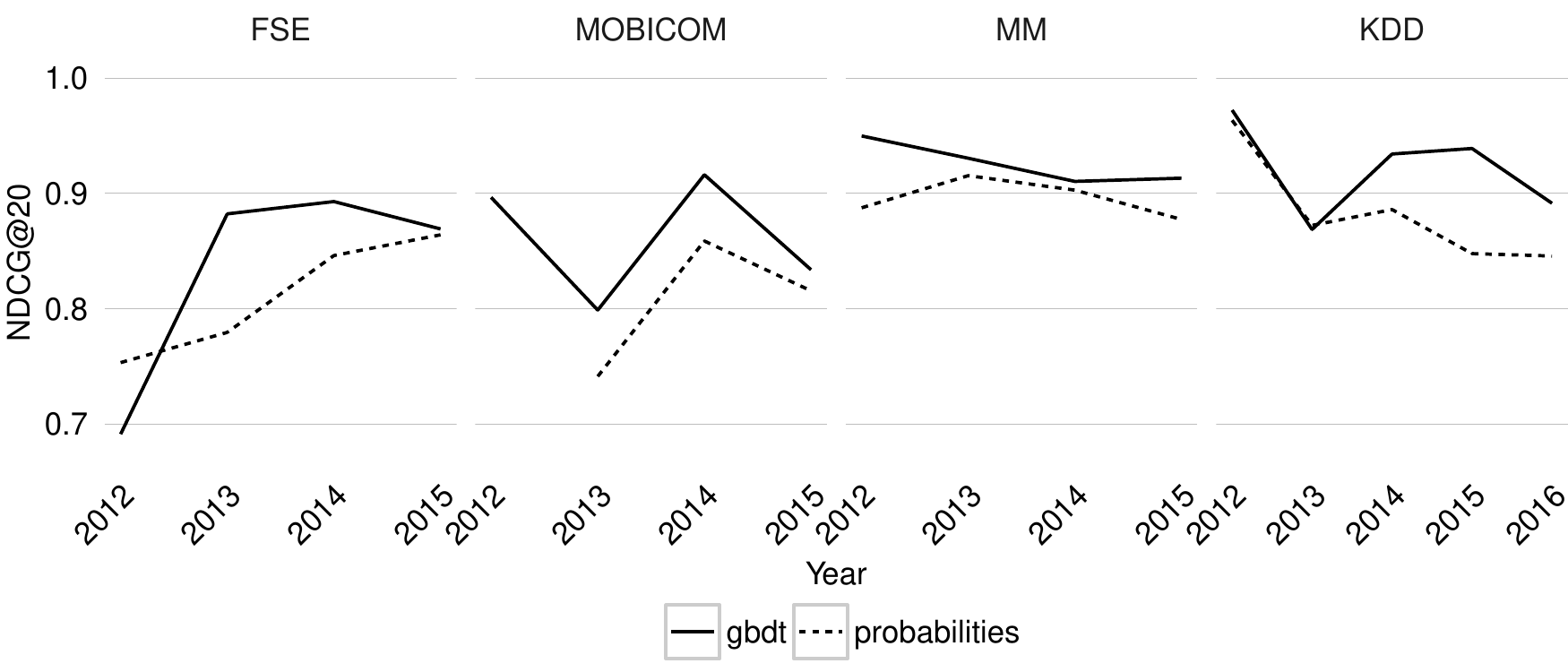} \caption[Results for the best features configuration for phase 3]{Results for the best features configuration for phase 3}\label{fig:benchmark_simple_probabilities}
\end{figure}

\end{knitrout}

\section{Discussions}
We have proposed and investigated several methods to rank the influence of affiliations at future conferences by predicting their number of accepted full research papers. Through our investigation across all phases of the competition we have observed the strong impact of the affiliations' past relevance. The short term influence trends as well as the longer term contributions of an affiliation to a conference are both strong predictors of the current relevance of the institution. 

We believe there are some underlying factors which have good predictive power for the relevance and this is what we have tried to capture in our models. There are of course other things which influence the affiliation's academic impact on the world, such as economic and business related factors. However we have not focused on these and only used the data provided by the competition organizers. We aimed to find a model which constantly improves with time, in the sense of being more accurate to predict the ranking for last year than it was at predicting the ranking for two years ago. The tuning of the features sets helped us choose the best models which at least constantly gave better results compared to the simple probabilities-based model.

Finally, we can think of a number of ways this work could be improved. One could be to experiment with different types of models using the same dataset or create an ensemble of the three models we have investigated. Another approach would be to explicitly model the degree of newcomers every year for a conference. Obviously, all these models would benefit from tapping into other data sources which can complement the information in the academic graph.

\bibliographystyle{abbrv}

\end{document}